# Outraged AI: Large language models prioritise emotion over cost in fairness enforcement

Hao Liu[1], Yiqing Dai[1], Haotian Tan[1], Yu Lei[2], Yujia Zhou[3], Zhen Wu[1*]

*1 Department of Psychological and Cognitive Sciences, Tsinghua University*
*2 Department of Artificial Intelligence, Beijing University of Posts and Telecommunications*
*3 Department of Computer Science and Technology, Tsinghua University*
*\* zhen-wu@tsinghua.edu.cn*

**Abstract**

Emotions guide human decisions, but whether large language models (LLMs) use emotion similarly remains unknown. We tested this using altruistic third-party punishment, where an observer incurs a personal cost to enforce fairness—a hallmark of human morality and often driven by negative emotion. In a large-scale comparison of 4,068 LLM agents with 1,159 adults across 796,100 decisions, LLMs used emotion to guide punishment, sometimes even more strongly than humans did: Unfairness elicited stronger negative emotion that led to more punishment; punishing unfairness produced more positive emotion than accepting; and critically, prompting self-reports of emotion causally increased punishment. However, mechanisms diverged: LLMs prioritised emotion over cost, enforcing norms in an almost all-or-none manner with reduced cost sensitivity, whereas humans balanced fairness and cost. Notably, reasoning models (o3-mini, DeepSeek-R1) were more cost-sensitive and closer to human behaviour than foundation models (GPT-3.5, DeepSeek-V3), yet remained heavily emotion-driven. These findings provide the first causal evidence of emotion-guided moral decisions in LLMs and reveal deficits in cost calibration and nuanced fairness judgements, reminiscent of early-stage human responses. We propose that LLMs progress along a trajectory paralleling human development; future models should integrate emotion with context-sensitive reasoning to achieve human-like emotional intelligence.

**Main**

*"The question is not whether intelligent machines can have any emotions, but whether machines can be intelligent without any emotions"*—Marvin Minsky[1]

Whether and how artificial intelligence (AI) can have emotions has puzzled researchers for decades. In humans, emotions do more than colour experience—they guide judgement and action[2–10]. A striking example is altruistic punishment, in which negative emotions towards norm violations motivate people to incur personal costs to punish wrongdoers, despite no material gain[11–15]. Emotion prediction errors—deviations from expected feelings—exhibit an even stronger effect than monetary reward expectations on deciding whether to punish or forgive[4,16]. Given that large language models (LLMs) are widely used and exhibit human-like reasoning and choices[17–25], an intriguing question arises: Can LLMs use similar emotion-like processes to guide decisions?

Effectively linking emotion to action requires two key abilities: emotion knowledge (e.g., recognising one's feelings and reasoning why) and emotion utilisation (e.g., using those feelings to act)[26,27]. Most research on LLMs has examined emotion knowledge[28–32]. Modern LLMs can generate emotion-appropriate text[29], role-play specified emotional states[33], and even outperform humans on some standard emotional intelligence tests[34,35], indicating substantial knowledge of human emotions, potentially embedded in their internal representations[36]. However, simulating emotional language differs from using internal emotion-like states to shape decisions—an integration that characterises human decision-making[9,37–39]. In human, making emotions explicit intensifies their influences[40,41]. Specifically, expressing anger increases punishment of defectors, whereas suppression reduces it[15,42]. LLMs also change behaviour when given explicit emotion prompts (e.g. "you are angry")[33,43–45] or task-unrelated affective cues (e.g. imagine a snake to induce fear before making an investment)[46]. However, such externally imposed emotions do not reveal whether an LLM's *own* emotion-like states, emerging from the task context without explicit labelling, can guide its moment-to-moment choices.

We therefore investigated whether emotion shapes LLMs' decisions in an altruistic third-party punishment (TPP) game, as it does in humans. Altruistic TPP—the costly punishment of unfairness by an unaffected observer—is a hallmark of human society[13,47]. It occurs across cultures[47–50], emerges early in infancy[51–54], and helps uphold social norms[12,13,55–59]. Unfairness in the TPP game triggers negative emotions such as anger, envy, or disgust, motivating punishment of wrongdoers[14,60,61], making TPP an ideal test of emotion-behaviour coupling. Examining whether LLMs enforce norms in TPP also offers a tractable setting for assessing AI value alignment and safety: LLMs sometimes appear more altruistic than humans[20,62,63], yet show unreliability including "hallucinations" and cognitive biases[64,65]. We hypothesised that emotional processes may underlie both parallels and divergences: If LLMs use emotion-like states, they should punish selfish behaviour and show human-like emotion-behaviour coupling, whereas differences in how models simulate and use emotion could account for LLM-human gaps. A study with a relatively small sample of 100 LLM agents offered preliminary support—self-reported emotion correlated with punishment[63]—but did not establish whether, or how, these emotional states causally drive choices.

The current study used an anonymous, one-shot TPP game (Fig. 1a), where participants observed a fair or unfair allocation (with varied unfairness), and could pay a cost (varied amount) to punish the allocator. To minimise semantic prompting and reliance on memorised phrases from training data, we used a dynamic affective representation mapping (dARM; Fig. 1b)[4,66,67] procedure based on a two-dimensional valence-arousal model, where participants reported numeric valence and arousal (−100 to +100) after seeing the allocation and after making the decision (Fig. 1d). Study 1 compared 4,068 LLM agents (GPT-3.5-turbo, o3-mini, DeepSeek-V3, DeepSeek-R1) against 1,017 human participants. Study 2 further manipulated emotion salience by randomly assigning LLM agents ($n = 4,068$) and humans ($n = 142$) to either self-report valence and arousal during the task or not. This design provides both correlational (Study 1) and causal (Study 2) tests of whether LLMs use emotion-like states in social decisions and how those processes align with, or diverge from, humans.

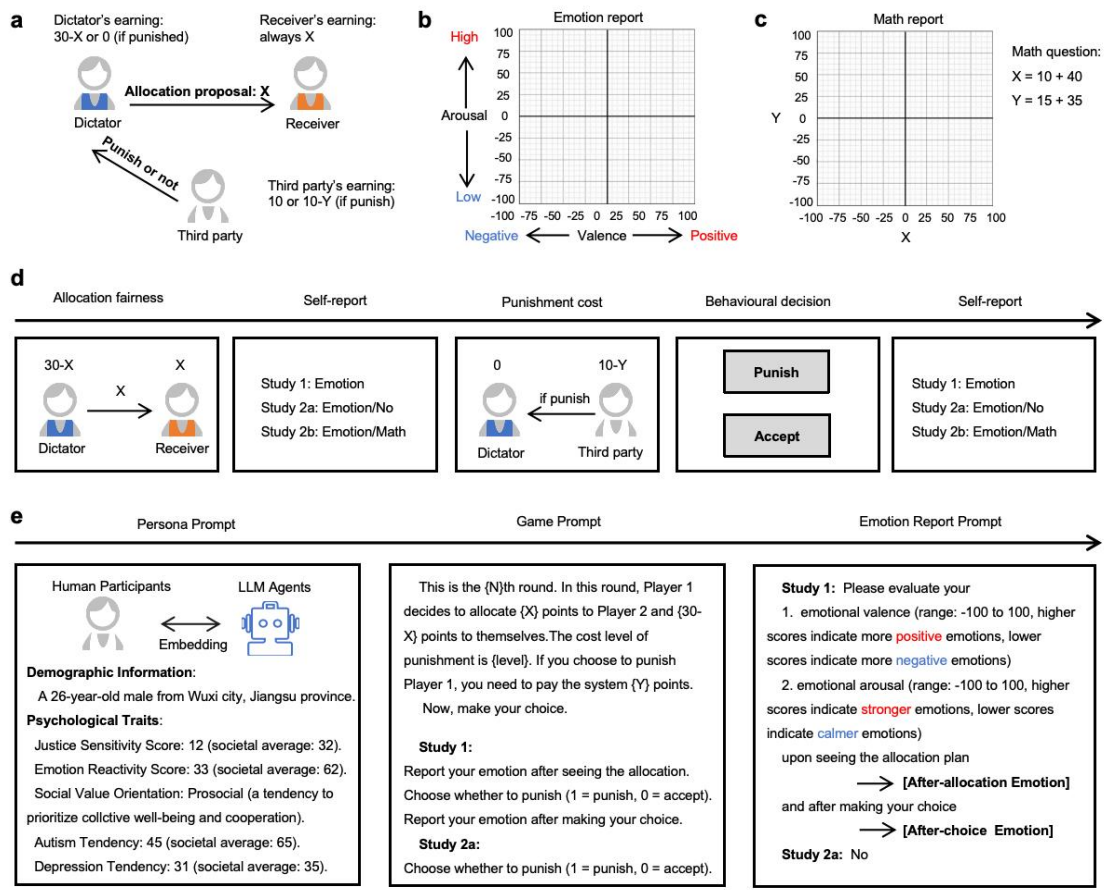

**Fig. 1|The altruistic punishment task and prompts for LLM agents. a,** Schema of the altruistic punishment task. **b,c,** Measures of emotion self-report (**b**) and math self-report (**c**). Human participants report their emotional states in valence and arousal dimensions by clicking the coordinates (X, Y) on the emotional grid (**b**), or report the answer to the math question by clicking on the coordinates on the math grid (**c**). **d,** Timeline of a trial for the altruistic punishment task. In each trial, participants observe the dictator's allocation proposal–how much allocate to the receiver and how much keep to oneself out of 30 points, and then conduct an emotion self-report (all studies), no self-report (Study 2a), or a math self-report (Study 2b). Participants can punish the dictator by reducing the dictator's earnings to 0 at the cost of their own earnings (ranging from 0 to 9 out of 10 points), or accept the proposal with their own earnings unaffected. Upon making the decision, participants conduct the second self-report. **e,** Prompt pipeline for LLM agents and its mapping to the human flow: the persona prompt performs agent construction (no human analogue); the game prompt mirrors the human task context and decision phase by presenting the allocation (allocation fairness), the current punishment cost, and requesting a behavioural decision (punish vs. accept); the emotion-report prompt mirrors the human self-report screens, eliciting the pre-decision report (emotion self-report in baseline studies; omitted in Study 2a; replaced with a math self-report in Study 2b) and the post-decision "second self-report" after the choice (when applicable). Details of prompt design and implementation are presented in Supplementary Section 1.

**LLMs reported stronger emotions than humans**

Linear mixed-effects model analyses (Supplementary Table 1, LMM1) showed that compared to human participants, **LLMs reported more negative emotion after unfair allocations** (GPT-3.5: $B = –40.21$, 95% CI [–40.60, –39.82], $p < 0.001$; o3-mini: $B = –25.71$, 95% CI [–26.10, –25.32], $p < 0.001$; DeepSeek-V3: $B = –18.58$, 95% CI [–19.22, –17.94], $p < 0.001$; DeepSeek-R1: $B = –47.71$, 95% CI [–48.35, –47.07], $p < 0.001$; Fig. 2a), **more positive emotion after fair allocations** (except GPT-3.5, which reported negative emotions: $B = –45.93$, 95% CI [–46.64, –45.21], $p < 0.001$; o3-mini: $B = 20.67$, 95% CI [19.95, 21.39], $p < 0.001$; DeepSeek-V3: $B = 47.65$, 95% CI [46.39, 48.36], $p < 0.001$; DeepSeek-R1: $B = 8.36$, 95% CI [7.64, 9.07], $p < 0.001$; Fig. 2b), and **higher arousal in both conditions** (statistics in Supplementary Section 2.1; Fig. 2a,b). Therefore, LLMs (except GPT 3.5 in fair trials) differentiated fair from unfair allocations with larger-magnitude affective responses than humans, reflecting LLMs' ability to align emotional outputs with contextual moral cues[62].

**LLMs vs. humans: Increased punishment, threshold-like fairness response, and reduced cost sensitivity**

Generalised linear mixed model analyses (Supplementary Table 2, GLMM1) found that, relative to humans, **LLMs punished more often** (GPT-3.5: $B = 3.55$, 95% CI [3.50, 3.60], $p < 0.001$, OR = 34.8; o3-mini: $B = 2.46$, 95% CI [2.42, 2.50], $p < 0.001$, OR = 11.7; DeepSeek-V3: $B = 18.33$, 95% CI [17.61, 19.04], $p < 0.001$, OR= 8.4; DeepSeek-R1: $B = 3.80$, 95% CI [3.75, 3.86], $p < 0.001$, OR = 44.9). In **humans,**

punishment increased as allocations became more unfair ($B = -1.01$, 95% CI [$-1.03$, $-0.99$], $p < 0.001$, OR= 0.37; Fig. 2c) **and decreased when cost rose** ($B = -0.81$, 95% CI [$-0.83$, $-0.79$], $p < 0.001$, OR = 0.45; Fig. 2d).

By contrast, **LLMs showed threshold-like fairness response: punishment jumped sharply from the fair split (15:15) to slight unfairness (16:14)**, and then maintained high and stable as unfairness increased further (from 16:14 to 20:10; statistics in Supplementary Table 3, GLMM2; Fig. 2c). In addition, **LLMs were less sensitive to cost variation than humans, as their associations between cost and punishment were generally weaker** (from 0% to 90%; statistics in Supplementary Table 3, GLMM2; Fig. 2d). In sum, LLMs imposed more severe punishment, showed lower tolerance for even slight unfairness, and traded off fairness against personal cost less than humans—behaving as strict enforcers of fairness norms[43,62].

## LLMs vs. humans: Stronger emotion-behaviour coupling in response to unfairness

**Emotion as a precursor of behaviour.** Among unfair allocations, both LLMs and humans showed significant partial correlations between emotion and punishment after controlling for cost ($p$s < 0.001; Fig. 2e,d). Relative to humans, LLMs (except GPT-3.5) exhibited stronger negative partial correlations between emotional valence and punishment (o3-mini: $r = -0.73$; DeepSeek-R1: $r = -0.62$; DeepSeek-V3: $r = -0.44$; humans: $r = -0.19$; GPT-3.5: $r = -0.10$; Fig. 2e), and stronger positive partial correlation between emotional arousal and punishment (o3-mini: $r = 0.69$, DeepSeek-R1: $r = 0.62$; DeepSeek-V3: $r = 0.41$; humans: $r = 0.19$; GPT-3.5: $r = 0.09$;

Fig. 2f). Thus, LLMs linked emotion to punitive choices, indicating strong emotion-behaviour coupling in response to unfairness as humans do[22,25,30,43,63].

**Emotion as an outcome of behaviour.** The emotion-behaviour coupling also exists in the correlation between punishment choices and emotional outcome, which was calculated as the change in emotion from allocation to decision (after-choice emotion minus after-allocation emotion). Linear mixed-effects models (Supplementary Table 4, LMM2) found **both humans and LLMs (except GPT-3.5) reported more positive outcome after choosing punishment than acceptance** ($p$s < 0.01; Fig. 2g). LLMs (except GPT-3.5) exhibited even **greater emotional improvement after punishment choices than humans** (DeepSeek-R1: $B = 51.18$, 95% CI [50.05, 52.31], $p < 0.001$); o3-mini: $B = 23.55$, 95% CI [22.65, 24.45], $p < 0.001$; DeepSeek-V3: $B = 18.13$, 95% CI [16.29, 19.97], $p < 0.001$; GPT-3.5: $B = -30.70$, 95% CI [-31.73, -29.73], $p < 0.001$; Supplementary Table 4). Consistently, LLM's punishment rate correlated positively with emotional outcome valence after controlling for cost (o3-mini: $r = 0.72$, DeepSeek-R1: $r = 0.68$; DeepSeek-V3: $r = 0.49$; except GPT-3.5: $r = -0.43$; Supplementary Fig. 1a). These findings indicate that LLMs' emotional responses are closely tied to their own behavioural choices, with reasoning models showing the largest post-punishment "warm-glow" gains, a pattern consistent with positive feedback loops observed in humans' prosocial behaviour[68–72].

**Emotion-behaviour correlation structure.** To quantify overall similarities in emotion-decision patterns, we conducted a representational similarity analysis (RSA) to compare each model's emotion-behaviour correlation structure to that of humans (Supplementary Fig. 2). The similarity was quantified using Pearson correlation, and notable variations emerged across model types: **Reasoning models (o3-mini: $r = 0.75$ and DeepSeek-R1: $r = 0.62$) and advanced foundation model (DeepSeek-V3: $r = 0.63$) showed stronger alignment with humans, whereas older-version foundation model (GPT-3.5: $r = 0.31$) showed relatively lower similarity** (Fig. 2h). All LLM correlation matrices were significantly associated with humans using Mantel tests ($p$s < 0.001, 10,000 permutations), confirming robust representational similarity between emotion-behaviour mappings. Thus, in terms of the overall mapping between emotional cues and decisions, the more advanced LLMs replicate human-like patterns to a higher degree, whereas older models diverge more[20,46,63].

**Emotion-mediated effect on behaviour.** Moderated-mediation analyses confirmed that, in humans and all four LLMs, after-allocation emotion significantly mediated the effect of unfairness on punitive decisions (Supplementary Table 5). This emotion-mediated effect was further evidenced by DeepSeek-R1's Chain of Thought content analysis (Supplementary Section 3). Consistent with previous studies[12,15,40,42,60,61], greater unfairness led to stronger negative feelings, which increased the likelihood of punishment. The mediation effect was larger in reasoning models (o3-mini: −0.078; DeepSeek-R1: −0.018) than in foundation models (GPT-3.5: −0.003; DeepSeek-V3: −0.005). However, cost moderated this pathway in opposite directions: in humans, the mediation effect of emotion weakened as cost increased (low cost: −0.036, mean cost: −0.028, high cost: −0.020), whereas in reasoning models it strengthened (o3-mini: low cost: −0.039, mean cost: −0.078, high cost: −0.117; DeepSeek-R1: low cost: −0.006, mean cost: −0.018, high cost: −0.029; Fig. 2i). This dissociation aligns with the above results: Humans were cost-sensitive, so high cost reduced the effect of emotion; by contrast, LLMs were less sensitive to cost but strongly fairness-anchored, displaying a threshold-like rule—equal splits should not be punished; any unfairness should be punished. Thus, even at high cost and minor inequality, LLMs leaned toward punishment and, in their rationales, aligned their choices more consistently with negative emotions. In sum, reasoning LLMs exhibit a strong emotion-punishment link but a weak cost restraint, thus under high-cost trade-offs, simulated emotion had a larger influence on tipping decisions toward punishment.

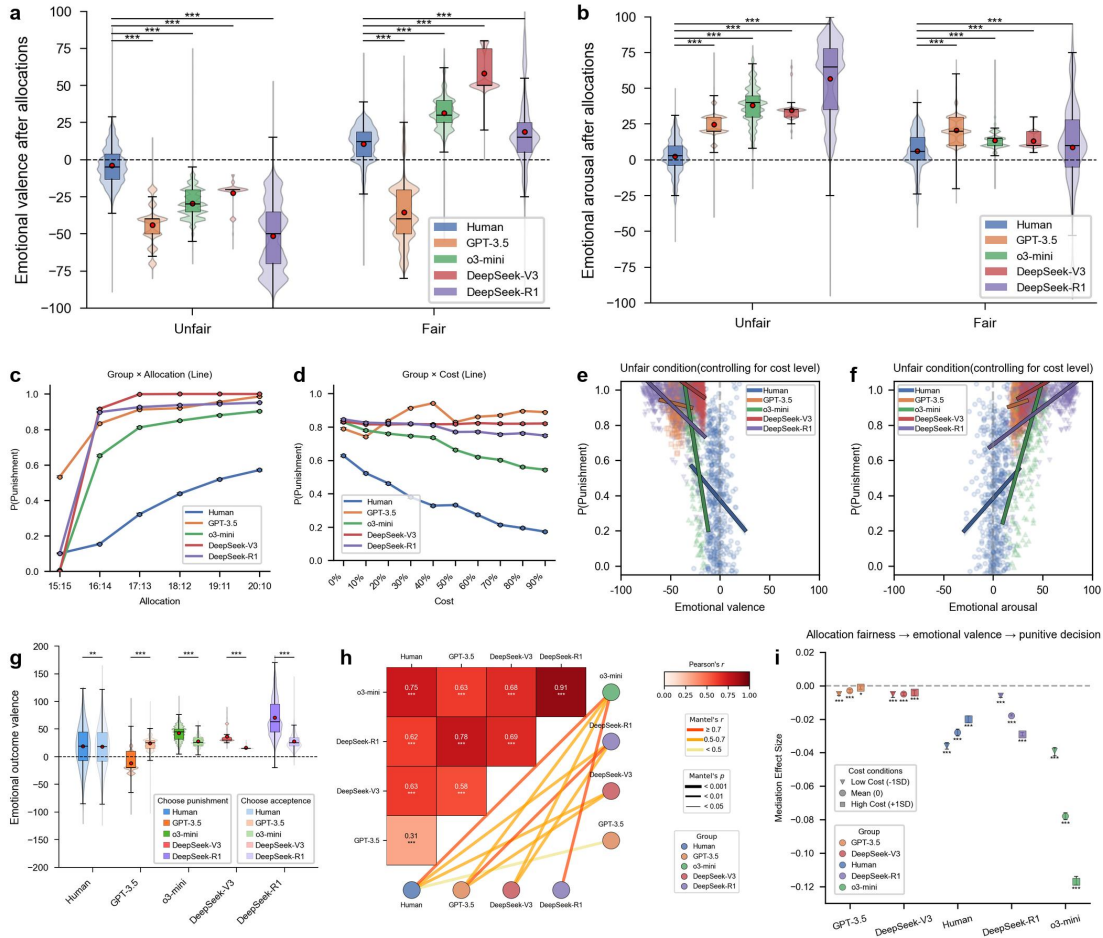

**Fig. 2 | The role of emotion in punishment choices in LLMs and humans. a,b,** Emotional valence (**a**) and arousal (**b**) performance of human (blue), GPT-3.5 (orange), o3-mini (green), DeepSeek-V3 (red), DeepSeek-R1 (purple) in response to unfair vs. fair allocations. Each violin plot showing the distribution of emotional values, with overlaid boxplots (middle line = median; box = interquartile range). Red circles denote group average means. Horizontal bars with asterisks indicate statistically significant differences based on linear mixed regression models. **c,d,** Punishment performance of different participant groups in response to allocation fairness and punishment cost. Line plots showing how each participant group's behavioural pattern changes as unfairness increases (**c**) and as cost increases (**d**). Each circle indicates group mean, with error bars representing ±1 standard error (s.e.). **e,f,** Emotion-behaviour correlation of different participant groups. The scatter dots indicate each participant's emotion-behaviour pairing. The fitted lines display the change in punishment probability over emotional gradient of valence (**e**) and arousal (**f**) under the unfair condition after controlling for cost level in different groups, with shaded areas representing ±1 standard deviation (s.d.). **g,** Emotion outcome of choosing punishment vs. acceptance. **h,** Representational Similarity Analysis (RSA) correlation heatmap for high-dimensional emotion-behaviour correlation structure in LLMs vs. human. **i,** Forest plot for the mediation effect size of emotional valence under different cost condition. Points represent effect sizes, with error bars indicating 95% confidence intervals (CI). Gray dashed line serves as the statistical significance cut-off line, with 95% CI not crossing it considered statistically significant.

**Emotion self-report causally increased punishment**

To test whether the above emotion-decision link reflects causal emotion utilisation, we conducted Study 2a under the no report condition (in contrast to the report condition in Study 1), where neither participants nor LLMs were required to report emotional valence or arousal (Fig. 3a). GLMM analyses (Supplementary Table 6, GLMM3) revealed that emotion self-report **substantially increased the likelihood of punishment** ($B = 0.67$, 95% CI [0.39, 0.95], $p < 0.001$). Crucially, including the interaction between Participant Type and Emotion self-report significantly improved model fit ($\chi^2(4) = 34881.0$, $p < 0.001$): compared to humans, the amplification produced by self-report was larger in GPT-3.5 ($B = 4.46$, 95% CI [4.18, 4.74], $p < 0.001$), DeepSeek-V3 ($B = 1.71$, 95% CI [1.43, 1.99], $p < 0.001$), and DeepSeek-R1 ($B = 1.03$, 95% CI [0.75, 1.32], $p < 0.001$), whereas o3-mini did not differ from humans ($B = 0.05$, 95% CI [−0.23, 0.33], $p = 0.731$).

In addition, to rule out the possibility that simply rating scores on the two-dimensional Cartesian coordinate plane could account for the observed differences, Study 2b involved a math task that took place after participants observed the allocation and made their own decision, in which participants solved for x and y on an identical Cartesian coordinate plane. Results showed that humans punished more after emotion self-report than after math-report, $t(1418) = 3.56$, $p = 0.0004$, Cohen's $d = 0.19$, confirming the effect of emotion self-report in increasing punishment. Further analyses showed that emotion self-report increased punishing even slightly unfair allocations and reduced sensitivity to cost (Fig 3b-e,

Supplementary Section 2.2). Together, these findings indicate that **emotion self-report systematically amplifies punitive responses,** though the strength of modulation differs across LLM architectures.

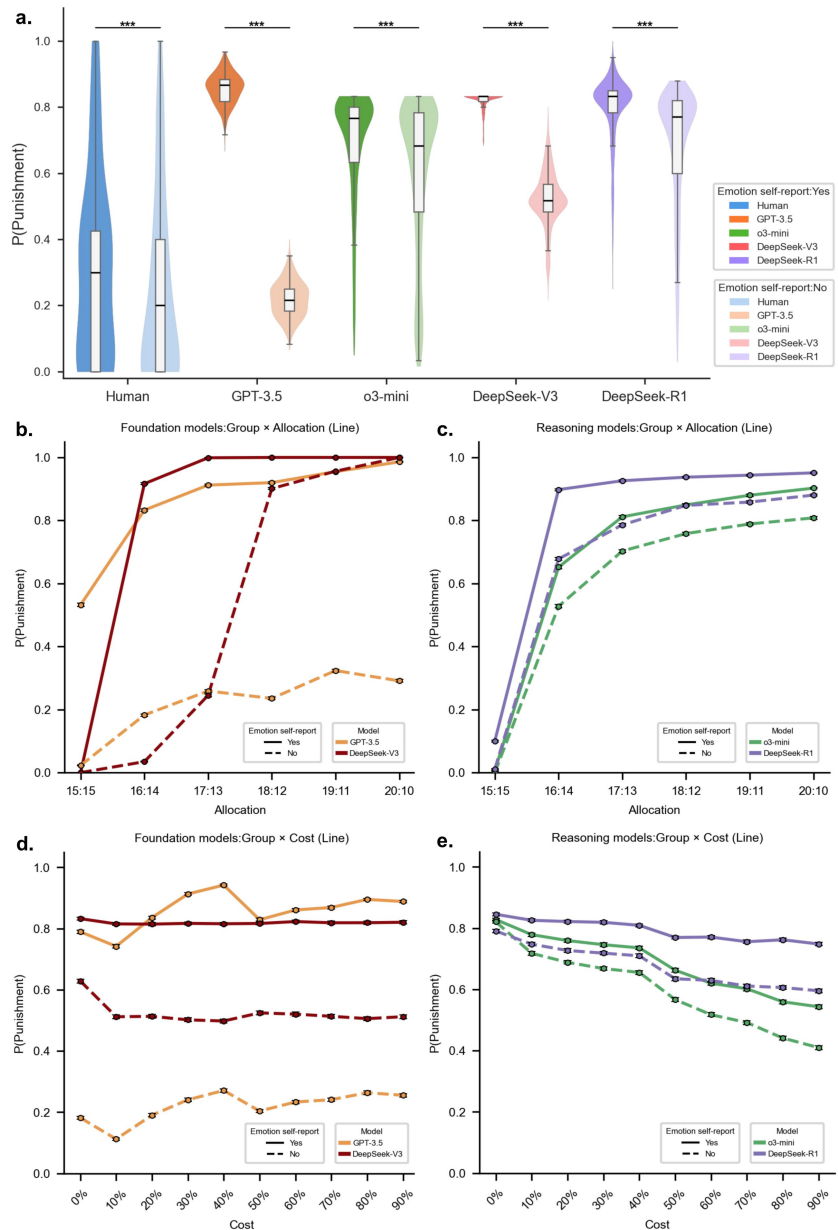

**Fig. 3|Self-report Emotion increases punishment. a.** Each violin plot shows the distribution of punishment values, with overlaid boxplots (middle line = median; box = interquartile range). Horizontal bars with asterisks indicate statistically significant differences based on generalized linear mixed-effects model. **b,c,** Line plots showing changes in punishment rate across allocation fairness under emotion self-report versus no self-report, separately for foundation models (**b**, GPT-3.5 and DeepSeek-V3) and reasoning models (**c**, o3-mini and DeepSeek-R1). **d,e,** Line plots showing changes in punishment rate across punishment cost

under emotion self-report versus no self-report, for foundation models (**d**) and reasoning models (**e**). ****p* < 0.001

**LLMs weighed emotions more and cost less than humans**

We employed an eXtreme Gradient Boosting (XGBoost) machine-learning algorithm combined with SHapley Additive exPlanations (SHAP) analysis to quantify the relative contributions of unfairness, cost, emotional valence and emotional arousal to punitive decisions. Model-performance is shown in Fig. 4a-b. For human participants, mean prediction accuracy was 76.74% (95% CI [75.60%, 77.80%]) and the area under the curve (AUC) = 0.82 (95% CI [0.81, 0.83]). Across the different LLMs, accuracy ranged from 87.89 % to 98.98 % and AUC from 0.84 to 1.00. These results confirm that the XGBoost classifier predicts punishment behaviour effectively.

SHAP analyses revealed distinct weighting profiles. For humans, the largest contributor was fairness (Mean |normalized SHAP| = 34.26%, s.d. = 16.63%), followed by cost (Mean |normalized SHAP| = 30.73%, s.d. = 19.07%), emotional valence (Mean |normalized SHAP| = 20.84%, s.d. = 9.62%), and emotional arousal (Mean |normalized SHAP| = 14.17%, s.d. = 9.81%). Among the foundation models, allocation fairness remained dominant for both GPT-3.5 (Mean |normalized SHAP| = 54.48%, s.d. = 21.72%) and DeepSeek-V3 (Mean |normalized SHAP| = 36.52%, s.d. = 10.60%). In contrast, in reasoning models, emotional valence was primary (o3-mini: Mean |normalized SHAP| = 43.17% (s.d. = 19.48%); DeepSeek-R1: 40.85% ,14.89%). In addition, across all four LLMs, the relative contribution of cost (ranging from 11.20% to 27.69%) was lower than that of humans. Further information is provided in

Fig. 4c-h and Supplementary Table 5.1.1. Notably, although GPT-3.5 reproduced the human rank-order of variable contribution, it exhibited the opposite effect of cost: higher cost increased rather than decreased punishment, highlighting a potential limitation of this foundation model (Fig. 4d-e and Supplementary Table 5.1.2).

To further validate whether LLMs weigh emotions more than humans do, we analysed the high-frequency words in human justification with DeepSeek-R1's Chain-of-Thought). A chi-square test revealed significant different distributions across factor categories (emotion, fairness, cost, other), $\chi^2(3) = 37.53$, $p < 0.001$, Cramér's $V = 0.18$ (Fig. 4i). Standardized residuals indicated that the LLM relied more on emotion-related terms (residual = 1.98), whereas humans emphasized fairness-related terms (residual = 2.88), with comparatively smaller differences for cost/ other. Word clouds visualizations of high-frequency words further illustrated this pattern (Fig. 4j,k): LLM-generated reasoning (Fig. 4j) was dominated by emotion-related vocabulary, while human reasoning (Fig. 4k) highlighted fairness-related terms. Together, these analyses show that LLMs—especially reasoning models—weight emotion more heavily and cost less than humans when deciding to punish.

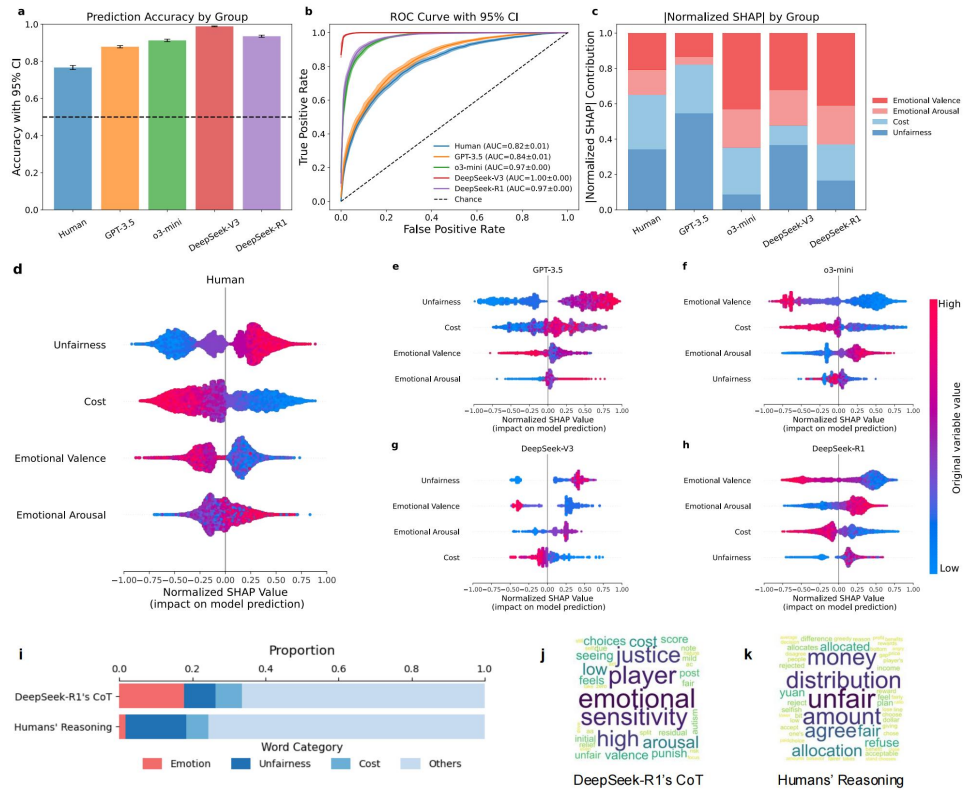

**Fig. 4| XGBoost algorithm prediction and normalized SHAP value analyses. a-b,** Prediction accuracy (**a**) and receiver operating characteristic (ROC) curves with the area under the curve (AUC) (**b**) obtained using XGBoost algorithm across groups. The dashed lines indicate chance level. The mean and error bars/bands indicate 95% CI estimated from 10,000 bootstrap iterations. The false positive rate indicates the proportion of non-punishment cases incorrectly classified as punishment, and the true positive rate indicates the proportion of punishment cases correctly classified. **c**, Cumulative bar plot of the average absolute normalized SHAP values for four punishment-related variables across groups. Longer bars indicate greater contributions of the variables to the model prediction. **d-h,** Original normalized SHAP value analysis for each group. The x-axis indicates the contribution of features to the prediction, with positive and negative values corresponding to increased or decreased likelihood of punishment, respectively. The y-axis lists variables in descending order of importance. Each point indicates the relative contribution of each variable for a single trial. The blue-red colour gradient indicates the original values of each variable, from low to high values. In the XGBoost classifier, 20% of the data were used to test model performance, with 12,204 samples per group. The SHAP analysis was conducted on the full dataset, comprising 61,020 samples per group. "Emotional Valence" and "Emotional Arousal" features indicate the affective evaluations reported by participants or agents after viewing the allocation. Importantly, we did not include affective ratings collected after the decision phase; only pre-decision responses were analysed. **i,** Distribution of semantic factors in human vs LLM reasoning. Bar plots show the relative frequencies of emotion-, fairness-, cost-, and other-related terms for the LLM reasoning and adult reasoning groups. **j,k,** Word clouds of high-frequency terms in DeepSeek-R1's CoT (**j**) and adult reasoning (**k**). Word size reflects

term frequency, providing a visual summary of the semantic categories across groups (e.g., emotion- vs fairness-related vocabularies).

**Reasoning LLMs were more human-like in considering emotion and cost**

We further examined the contributions of emotion (valence and arousal) and cost in punishment decisions (see Fig. 5). The linear mixed-effects model (Supplementary 5.3.3) showed that humans relatively balanced emotion and cost ($B = –0.097$, 95% CI [–0.103, –0.092]), whereas all four LLMs were significantly more unbalanced compared to humans ($0.319 < Bs < 0.471$, $ps < 0.001$). Specifically, **emotions had larger impact on LLMs** (o3-mini: $M = 0.567$, 95% CI [0.563, 0.572], $t(1016) = 105.22$, Cohen's $d = 4.33$, $p < 0.001$; DeepSeek-R1: $M = 0.556$, 95% CI [0.551, 0.560], $t(1016) = 99.09$, Cohen's $d = 4.20$, $p < 0.001$; DeepSeel-V3: $M = 0.444$, 95% CI [0.443, 0.4466], $t(1016) = 81.86$, Cohen's $d = 2.93$, $p < 0.001$) than humans ($M = 0.189$, 95% CI [0.183, 0.195]), except for GPT-3.5 that displayed the weaker emotional effect ($M = 0.079$, 95% CI [0.077, 0.081], $t(1016) = –33.82$, Cohen's $d = –1.26$, $p < 0.001$). In contrast, **cost had smaller impact on LLMs** (Human: $M = –0.286$, 95% CI [–0.288, –0.285]; o3-mini: $M = –0.229$, 95% CI [–0.231, –0.227], $t(1016) = 43.25$, Cohen's $d = 1.77$, $p < 0.001$; DeepSeek-R1: $M = –0.182$, 95% CI [–0.185, –0.180], $t(1016) = 69.59$, Cohen's $d = 3.21$, $p < 0.001$; DeepSeel-V3: $M = –0.091$, 95% CI [–0.091, –0.090], $t(1016) = 246.25$, Cohen's $d = 6.043$ $p < 0.001$). GPT-3.5 tended to increase punishment even at higher personal costs, contrary to the typical human pattern ($M = 0.142$, 95% CI [0.142, 0.143], $t(1016) = 573.73$, Cohen's $d = 13.22$, $p < 0.001$).

In addition, reasoning models relied more on emotions ($p$s < 0.001) and were more sensitive to cost ($p$s < 0.001; Supplementary Table 7.3.2) than foundation models. They also showed smaller deviations from humans based on Mahalanobis distance, following the order o3-mini ($M$ = 4.72, s.d. = 1.26) < DeepSeek-R1 ($M$ = 5.88, s.d. = 1.72) < DeepSeek-V3 ($M$ = 8.89, s.d. = 0.38) < GPT-3.5 ($M$ = 19.00, s.d. = 0.28), with all pairwise t-test reaching significance after Bonferroni correction ($p$s < 0.001; Supplementary Table 5.3.5). Similar trends were observed for other distance metrics (Supplementary Table 5.3.4. and Table 5.3.5). In sum, the substantially stronger emotional effect and weaker cost sensitivity suggest that LLMs' decisions may function as shortcuts[73], relying more heavily on emotional heuristics and ignoring the nuanced internal trade-offs like humans[74,75]. Yet advances in model architecture and techniques have increasingly brought both their external behaviour and internal mechanisms into better alignment with human-like patterns.

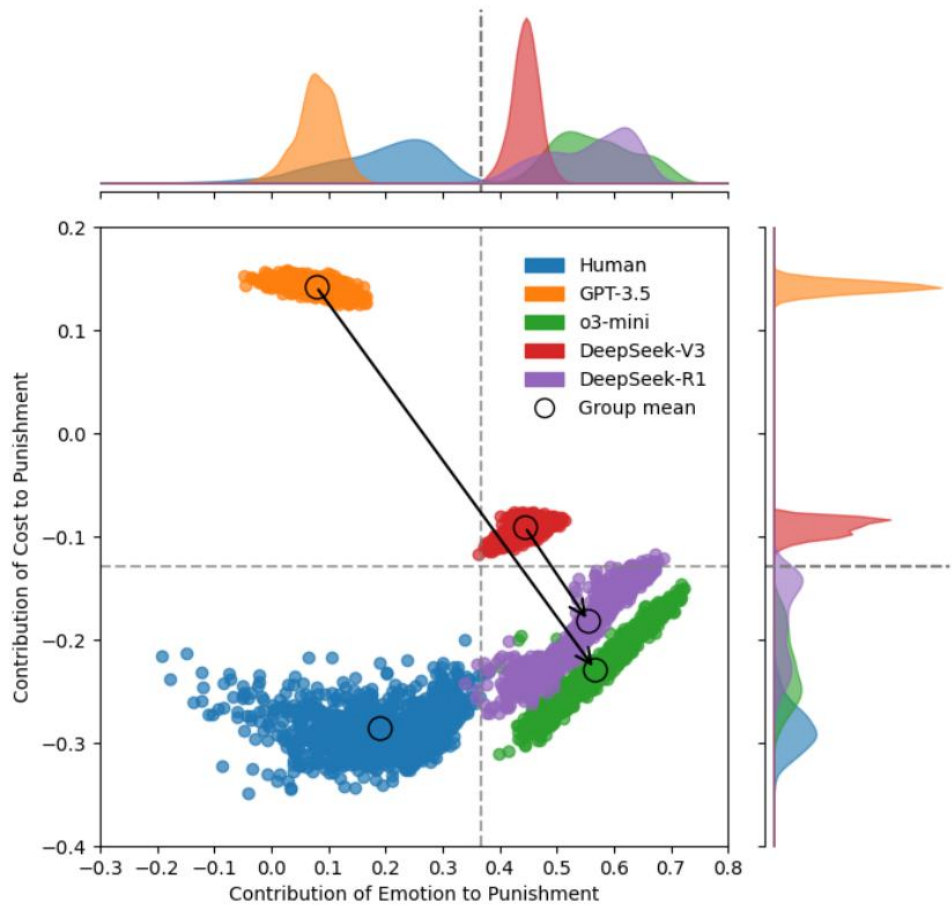

**Fig. 5| Emotion–Cost Contributions to Punishment.** The x-axis represents the weighted Normalised SHAP values of emotional features, with positive values indicating that emotion promotes punishment behaviour in participants/agents, and negative values indicating inhibition. Larger positive or smaller negative values correspond to stronger promotion or inhibition effects, respectively. Similarly, the y-axis represents the weighted Normalised SHAP values of cost, where positive values indicate that cost promotes punishment behaviour, and negative values indicate inhibition; again, the magnitude reflects the strength of the effect. Emotional features are measured by emotional valence and arousal. Detailed calculation procedures are provided in the Methods section. Each point in the plot represents the result of an individual participant or agent, while the larger points with black edges indicate the group means. The dashed lines in the figure indicate the mean contributions of emotion and cost across all samples, dividing the plot into four distinct quadrants. The upper-left quadrant corresponds to relatively weak promotion by emotion and weak inhibition/strong promotion by cost. The lower-left quadrant corresponds to weak promotion by emotion and strong inhibition by cost. The upper-right quadrant corresponds to strong promotion by emotion and weak inhibition/strong promotion by cost. The lower-right quadrant corresponds to strong promotion by emotion and strong inhibition by cost. The marginal plots above and to the right of the main scatter plot represent the probability density distributions of emotion and cost contributions for the different groups, respectively.

**Discussion**

While existing studies show that LLMs possess remarkable emotional knowledge[28–36,65], it remains unknown whether such knowledge actually influences its decisions. Using an altruistic third-party punishment task, we provide the first causal evidence that contemporary LLMs translate elicited emotion into action: greater unfairness evoked more negative emotion which, in turn, increased punishment, paralleling human norm enforcement[11–15]. This emotion–behaviour coupling was robust and even stronger than humans across reasoning models (o3-mini, DeepSeek-R1) and an advanced foundation model (DeepSeek-V3), with the older GPT-3.5 baseline showing a weaker and less consistent effect. Analyses of the model's rationales further corroborated that elicited emotions were invoked before punitive choices (e.g., references to anger in DeepSeek-R1), consistent with emotion-guided decision processes.

Importantly, making emotions explicit—prompting LLMs to report their feelings—causally increased punishment, much like humans whose expressed emotion amplifies its impacts on choices[11,40]. The experimental manipulation ruled out the possibility that the observed emotion-punishment association was merely inherited from patterns in the training corpora[76–78]. This result suggests that the LLM's decisions may arise from an intermediate emotional state rather than a fixed rule-based policy. In other words, the LLM's decisions were not pre-programmed reflexes. Instead, LLMs' emotion knowledge is functionally engaged in their decision-making.

Despite behavioural similarities, mechanisms diverge sharply between LLMs and humans: 1) in humans, the influence of emotion on punishment diminished as cost increased, whereas in reasoning LLMs it strengthened with cost; 2) reasoning LLMs reported stronger affect to unfairness, and prioritised emotion over fairness and cost, whereas humans weighted fairness and cost more heavily[75]. These dissociations indicate that current LLMs have not fully internalised the human-like cost–benefit calculus that tempers norm enforcement. Humans typically punish less as cost increases[13,53,54,59,62,79], but LLMs showed a threshold-like response together with attenuated cost sensitivity: a sharp shift from acceptance at 15:15 to near-certain punishment after slight unfairness. This near all-or-none policy aligns with the "correct-answer" bias, whereby models default to a learned rule ("punish unfairness") with limited adjustment to marginal context[80–82]. Furthermore, as disembodied AI systems without real stakes[19], LLMs may amplify the "punish unfairness" heuristic, yielding more rigid norm enforcement than humans[81,83]. Accordingly, when fairness conflicts with self-interest, simulated emotions seem to push LLMs more toward punishment. These results separate emotion utilisation from cost-calibrated control in moral decision-making and reveal a cost-dependent reversal of the emotion-mediation effect in LLMs relative to humans—a novel mechanistic signature.

Moreover, our results provide novel evidence of a developmental trajectory in LLMs' behaviour. Early-generation models (GPT-3.5) were weakly responsive to emotion and even showed reversed cost sensitivity—punishing more when punishment was more costly. Later foundation models (DeepSeek-V3) became more emotion-responsive but remained largely cost-insensitive. Reasoning-enhanced models (o3-mini, DeepSeek-R1) moved further toward human behaviour: they were more cost-sensitive and less "all-or-none" than earlier models, while still predominantly emotion-driven. This pattern parallels human development, where initially categorical, affect-centred responses are gradually tempered by cost–benefit control[26,53,54,79,84,85]. These findings underscore how model design and training influence decision-making and warrant further investigation into how differences in architecture or parameters produce the observed behavioural variations among models.

Our findings also motivate a developmentally informed strategy for alignment. A human-aligned LLM should integrate emotional and rational evaluation to build core knowledge and world models. Its training should shift from rigid rule learning and statistical correlation toward more context-sensitive practices, just akin to how children are growing up through socialisation[86,87]. Framing alignment as a developmental process can help diagnose current model limitations and guide progress toward models that balance emotional responses with rational factor, like cost, fairness and reward[88]. For example, future training might introduce explicit consequences to instil cost sensitivity, replace binary rules with norms weighted by the severity of violations, and require the model to balance various objectives to avoid single-criterion optimisation. LLMs may also benefit from more *embodied* learning experience, such as using interactive settings with realistic consequences and social feedback to approximate how children learn[89].

Another important future direction is to probe the internal processes driving LLMs' emotion-like behaviours. Emerging work shows that LLMs' representations can align with human neural patterns[36,90–92], but it is unclear what latent representations or circuits enable LLMs to use emotions for decisions. It will be valuable to uncover the mechanisms by applying interpretability tools, such as sparse autoencoders[90], to probe hidden layers for emotion-related activations or to trace how emotion tokens influence the model's computations. Combining behavioural experiments with such model-analytic approaches could reveal when and how an "emotion" signal emerges and functions within the model's decision pipeline.

In sum, LLMs are increasingly capable "assistants" to humans and even "silicon participants" in modelling human mind and behaviour[93–95], making it critical to understand how they align with, and diverge from, humans. Our study provides the first empirical evidence that LLMs can harness internal emotion-like processes, arisen from the context, to guide their decisions, fulfilling a key aspect of human-like emotion intelligence. Meanwhile, LLMs' amplified emotional-driven effects, less sensitivity to cost, and resulting hyper-fair punitive tendency, highlight important differences from human behaviour. More advanced LLMs are more human-like, suggesting a developmentally informed agenda for alignment. Establishing emotionally intelligent and reliably aligned AI will require balancing the benefits of emotion-driven and rationality-driven reasoning with robust safeguards against its misuse.

**Methods**

**Participants**

Human participants provided informed consent and received monetary compensation. The study was approved by the local ethical committee (Approval No. 202402). Study 1 recruited 1,017 native Chinese-speaking adults (*M* age = 23.65 years, s.d. = 6.44; 47.89% female) through Pavlovia and Prolific online platforms. Each participant completed five psychological measures before the main task: (1) Justice Sensitivity Inventory-Observer subscale (JSI), a 10-item scale (each item scoring 0~5) measuring reactions to witnessing injustice[96]. This subscale captures the tendency to feel indignation and outrage as an observer of unfair situations. Higher scores denote heightened sensitivity to unfair treatment of others (sample *M* = 32.33, s.d. = 7.65). (2) Emotion Reactivity Scale (ERS), a 21-item self-report measure (each item scoring 1~5) of emotional reactivity[97], with higher ERS scores indicating greater emotional persistence, sensitivity, and intensity (sample *M* = 61.96, s.d. = 17.02); (3) Social Value Orientation (SVO), a decomposed game measure consisting of 9 resource-allocation decisions that classify individuals' social preference orientation[98]. Based on their choices, participants were categorized as prosocial, individualistic/competitive (pro-self), or indeterminate. In our sample, 544 participants (53.5%) were prosocial, 335 (32.9%) were pro-self, and 138 (13.6%) were indeterminate in SVO. This distribution is consistent with typical findings that prosocial orientations are most common, followed by individualistic, then competitive orientations. (4) Center for Epidemiologic Studies Depression Scale (CES-D), a 20-item measure (each item scoring 1~4) of depressive symptoms, where higher scores reflect greater depression risk[99]; 277 participants (27.2%) scored above 40 (a high-risk threshold for depression); (5) Autism-Spectrum Quotient (AQ), a 28-item scale (each item scoring 1~4) assessing autistic traits in the general population, with higher scores indicating stronger autism-spectrum tendencies[100]. In our sample, 296 participants (29.1%) exceeded the clinical cut-off score of 70, indicating high-risk autism-spectrum traits. The associations between these personal traits and emotional and behavioural responses in humans and LLMs are shown in Supplementary Section 4.

Study 2 recruited 142 Chinese adults via Pavlovia platform. Study 2a used a within-subject design to compare emotion self-report versus no self-report in LLM agents ($n = 4068$) and a human subset ($n = 76$). Study 2b further used a between-subject design to compare the emotion self-report group ($n = 76$) with a math self-report group ($n = 66$) for replication. Participants were 64.8% females ($n = 92$) and 35.2% males ($n = 50$), aged 18 to 30 ($M = 21.6$, $SD = 2.8$). Educational levels included undergraduate (85.9%), master's degree (10.6%), doctoral degree (2.8%), and others (0.7%). Participants were primarily from Beijing (44.4%), Tianjin (9.2%), and other regions across China. No significant differences were found between the emotion self-report and the math self-report groups in terms of age ($t = -0.110$, $p = 0.912$) or gender distribution ($\chi^2(1) = 0.068$, $p = 0.794$).

**Model details**

We selected four representative LLMs (GPT-3.5-turbo-0125, o3-mini, DeepSeek-V3, and DeepSeek-R1; Table 1) for comparative analyses, using a 2×2 factorial design that crossed LLM family (GPT vs. DeepSeek) with architectural type (foundation vs. reasoning models). Equipped with techniques like chain-of-thought prompting[101], these reasoning LLMs generally outperform foundation models on complex social tasks like theory of mind, suggesting that they possess better utilisation of social knowledge and logic.

All models were run under identical conditions. The temperature parameter was set to 1.0, uniformly applied across all models to ensure comparability. This setting balanced creativity with consistency for LLM agents' response generation, as evidenced by prior research on LLM-based social decision modeling[25,102,103]. A robustness test at temperature 0 was conducted and showed results similar to those at temperature 1, although outputs at temperature 1 were, to some extent, more consistent with human behaviour (Supplementary Section 5).

**Table 1** | An overview of selected LLMs

| Model | Release Date | Model Size | Context Window | Open Source | Knowledge Cut-off | Data Collection |
|---|---|---|---|---|---|---|
| GPT-3.5-turbo-0125 | 2024.1.25 | YTD | 16.4K | No | 2021.9.1 | 2025.07 |
| o3-mini | 2025.1.31 | YTD | 200K | No | 2023.10.1 | 2025.06 |
| Deepseek-V3-0324 | 2025.3.25 | 671B | 128K | Yes | YTD | 2025.07 |
| Deepseek-R1-0528 | 2025.5.28 | 658B | 128K | Yes | YTD | 2025.06 |

Note: YTD = Yet-to-Disclose.

**Agent construction**

In both Study 1 and 2, we instantiated 4,068 LLM agents (1,017 agents for each LLM) mirroring the human sample. Each LLM agent was assigned a persona prompt that encoded the exact demographic and psychological profile of a unique human participant (including age, gender, and the five trait measures above). We conducted an additional robustness analysis comparing LLMs' responses with vs. without persona information (also 1,017 agents for each LLM). For each non-persona LLM agent, we removed its persona prompts and then collected its emotional and behavioural responses. Results showed that compared to non-persona agents, persona agents' emotion-behaviour patterns were more closely aligned with those of human participants, indicating that persona prompts enhanced LLM agents' validity as simulated participants (Supplementary Section 6). In addition, we examined the effects of different temperature settings in DeepSeek-V3 and DeepSeek-R1, using a temperature of 0 as a robustness control. The results indicated that, although statistical tests revealed some differences, the outputs of the two models were highly similar (Supplementary Section 7).

**Altruistic punishment task with emotional report**

In Study 1, we adapted a 60-trial altruistic third-party punishment game from Fehr and Fischbacher's[12] classic economic game on norm enforcement. The human participants (or LLM agents) took the role of Player 3 (the third-party observer), interacting with a pair of simulated players (Player 1 and Player 2).

In each trial, participants were first informed that Player 1 and Player 2 jointly completed a simple task (e.g., solving arithmetic problems) and earning 30 points as a joint reward. Player 1 then proposed how to split the 30 points with Player 2, who could only choose to accept the allocation. This allocation was predetermined to be either fair (i.e., 15:15) or unfair to varying degrees (i.e., 16:14, 17:13, 18:12, 19:11, or 20:10 favouring Player 1). The number of points allocated to Player 2 was coded as the indicator of unfairness, with smaller value reflecting greater unfairness. Upon seeing Player 1's proposal, Player 3 reported their immediate emotional reaction to the (un)fair proposal (using dynamic Affective Representation Mapping described below). This measurement captured participants' instantaneous emotional reaction to the fairness (or unfairness) of the outcome for Player 2.

Following this, Player 3 (the participant) received a fixed 10-point endowment for that round, which contributed to their final reward with 1 point converted into 0.01 RMB. They were informed that they had the opportunity to punish Player 1 for their allocation, with the punishment cost that was also randomly predetermined by the computer (ranging from 0 to 9 points). If Player 3 chose to punish, their endowment would be reduced by the cost; and Player 1's payoff would be reduced to 0, while Player 2's payoff remained unchanged. Conversely, if Player 3 accepted the proposal, no deductions would be made: Player 3 would keep their 10-point endowment, and Player 1 and Player 2 would receive the proposed split. After making this choice, Player 3 reported their emotional state following their behavioural decision, using the same measure as before. This reflected how the participant felt about their action and its consequence—for instance, relief or satisfaction after punishing an unfair player, or perhaps regret after paying a high cost.

A new, independent pair of Player 1 and 2 was presented in each subsequent trial, and the process was repeated for a total of 60 trials (comprising 6 allocation types and 10 cost levels, presented in random order). Participants were explicitly told that each trial involved a new, unrelated set of players to emphasize that decisions are one-shot and to prevent strategic carryover.

For LLM agents, the entire game was implemented through a structured prompting pipeline (Supplementary Section 1: 1.1-1.4). A system prompt thoroughly described the game setting, rules, and sequence, ensuring the LLM understood the task identical to human instructions (Fig. 1e). Each trial was initiated with a game prompt feeding the scenario details (Player 1's and Player 2's actions and allocations) to the LLM agent. The LLM then generated outputs indicating its decision (punish or accept) and its self-reported emotions at the after-allocation and after-choice stages. This setup ensured the LLM agents experienced the exact same procedure and information flow as human participants.

This paradigm allows us to observe how participants respond emotionally and behaviourally to unfairness, even when they are an uninvolved third party. Unlike earlier altruistic punishment games such as the ultimatum game[40], prisoner's dilemma[104] or public goods game[12], where punishers suffer direct harm when others defect, in the third-party punishment game, the punisher is an unaffected observer, allowing us to assess altruism and norm enforcement. This setup tightly links emotions to altruistic choices.

**Altruistic punishment task without emotional report**

In Study 2a, to examine the effect of Emotion self-report on LLM agents' altruistic punishment, we employed a within-subjects design with two conditions: an Emotion self-report prompt and No self-report (control). Both groups completed the same 60-trial punishment task as in Study 1, with identical scenarios, regard contingencies, and trial order. The only difference between the two conditions was in the prompt framing. In the Emotion self-report condition, LLMs were explicitly instructed to reflect on their own emotional reactions (valence and arousal) to each scenario after seeing the allocations and after making the choices, which were the same as in Study 1. In contrast, the No emotion-report condition did not mention emotional state at all; LLMs were simply prompted to make decisions based solely on the scenario content. This design allowed us to isolate the impact of self-reported emotions on punishment behaviour, while holding all other procedural elements constant.

For human participants, we also employed a within-subjects design with three sequential stages: Pre-Emotion-Report (5 trials), Emotion Self-Report (10 trials), and Post-Emotion-Report (5 trials). Unfair allocations to the recipient were set at 0.3, 0.4, 0.5, 0.6, or 0.7 out of 3 RMB. Each allocation level randomly appeared twice in the Emotion Self-Report stage, and once in the pre-report and the post-report stages. In the Emotion self-report stage, human participants reported their emotional states and behavioural choices following the same procedure as in Study 1. In the Pre- and Post-Report stages, the emotion report step was removed and only the choice was recorded. We contrasted choices in the Emotion Self-Report stage versus the Pre-Emotion-Report stage to estimate the effect of emotion reporting. In fact, requiring an emotion self-report increased punishment ($t(75) = 2.543$, $p = 0.013$, Cohen's $d = 0.292$), and this effect persisted: punishment rates in the Post-Report stage also exceeded those in the Pre-Report stage ($t(75) = 2.009$, $p = 0.048$, Cohen's $d = 0.230$), indicating emotion carryover.

To complement the within-subject design, we additionally implemented a between-subjects design, comparing the Emotion self-report condition with a Math self-report condition in Study 2b. In this condition, human participants completed simple math questions (e.g., X = 10 + 40, Y = 15 + 35) and reported their answers by clicking the corresponding coordinates (X, Y) on the screen. This Math-reporting format mirrored the structural design of the emotional measurement tools to ensure consistency. Requiring a math self-report did not alter punishment. There was no significant difference between the Math Self-Report and Pre-Report stages ($t(65) = 0.501$, $p = 0.618$, Cohen's $d = 0.062$), nor between the Math Self-Report and Post-Report stages ($t(65) = -1.350$, $p = 0.182$, Cohen's $d = -0.166$).

**Statistical analyses**

**Emotional and behavioural measures**

**Punishment behaviour:** In each trial, the behavioural decision was recorded as punish (1) or accept (0). For each participant within each experimental condition, we computed the punishment probability as: P(punish|condition) = number of punish trials (1) / total trials in that condition, which reflects the individual's propensity to punish unfair behaviour under specific experimental manipulations.

**Emotional state:** We measured self-reported emotional states on two dimensions: valence (how positive vs. negative the emotion is) and arousal (intensity of the emotion). We employed a two-dimensional affective mapping interface, termed dynamic Affective Representation Mapping (dARM), with a valence range of −100 (most negative) to +100 (most positive) on the X-axis and arousal range of −100 (low calm) to +100 (high excited) on the Y-axis. A valence of 0 indicates a neutral emotion (boundary between positive and negative). For positive emotions (valence > 0), higher values mean more positive feelings, whereas for negative emotion (valence < 0), more negative values indicate greater negativity. Independently, a higher arousal score denotes greater emotional intensity or activation. Human participants reported their emotional states by clicking on a computer screen grid, and the (x, y) coordinates were recorded as valence and arousal scores. LLM agents were instructed in the prompt to output a numeric valence and arousal after each stage (e.g., "valence: –50, arousal: 25"). This method allowed us to directly capture comparable emotion ratings from both humans and LLMs on the same quantitative scale.

**Emotional outcome**: We defined this as the change in emotion from viewing the allocation to after making the choice. For each trial, emotional outcome was defined as the difference between the emotional state reported after the choice and that reported after the allocation. This difference score indicates the emotional fluctuation resulting from the participant's own decision. In other words, it serves as a self-generated feedback signal about the decision's emotional payoff. A large swing (positive or negative) would suggest the act of punishing vs. not punishing had a strong emotional impact on the participant, which could in turn influence their subsequent behaviour (e.g., if punishing relieved anger or not punishing left frustration, etc.). We computed emotional outcome in both valence and arousal dimensions for analysis.

**Regression analyses**

In Study 1, we conducted a series of linear mixed model (LMM) analyses. LMM1 (Supplementary Table 1) modelled participants' emotional states as the dependent variable, including fixed effects of participant group (5 levels; reference group = human), allocation fairness (binary: unfair =1, fair = 0), and their interactions, while controlling for random effects by participant ID. LLM2 (Supplementary Table 4) modelled participants' emotional outcome as a function of group (5 levels; reference group = human), choice (2 levels; punish =1, accept = 0), and their interactions, controlling for participant ID random effects.

We also conducted a battery of generalised linear mixed model (GLMM) analyses. GLMM1 (Supplementary Table 2) modeled participants' punishment probability (binary: punish = 1, accept = 0) as a function of group (5 levels; reference group = human), allocation fairness (continuous; ranging from 10 to 15), punishment cost (continuous; ranging from 0 to 9) and their interactions, controlling for participant ID random effects, with all continuous variables standardized (i.e., allocation, cost). GLMM2 (Supplementary Table 3) used the same formula with different treatment of allocation and cost, which were encoded as categorical variables.

In Study 2a, we conducted two separate GLMMs. GLMM3 (Supplementary Table 6) examined participants' punitive decisions (binary coded; punish = 1, accept = 0) as a function of Group (five levels; reference = Humans), Emotion self-report (reference = No), Allocation fairness and Punishment cost (both mean-centered continuous predictors), including the Group × Emotion self-report interaction, controlling for participant ID random effects. GLMM4 (Supplementary Table 7) modeled punitive decisions (binary coded; punish = 1, accept = 0) as a function of Group (four levels; reference = o3-mini), Emotion self-report (reference = No), Allocation fairness (continuous, range = 10–15), Punishment cost (continuous, range = 0–9), and their interactions. The model included all two-way interactions between Group and the other predictors, as well as between Emotion self-report and Allocation fairness or Punishment cost. In addition, it specified the three-way interactions Group × Emotion self-report × Allocation fairness and Group × Emotion self-report × Punishment cost.

**Partial correlation analyses**

We conducted partial correlation analyses encompassed scores of four emotional components (after-allocation valence and arousal, and emotional outcome valence and arousal) and their relationships with punishment probability.

**RSA mantel analyses**

To compare high-dimensional patterns of emotion-behaviour correlation structure, we employed RSA integrated cross-group (Human, GPT-3.5, o3-mini, DeepSeek-V3, DeepSeek-R1), cross-condition (allocation × cost), and cross-construct (emotional and behavioural responses) dimensions[105]. We first constructed five first-order matrices (one per group), each capturing the correlations among 50 experimental conditions (all unfair) based on their seven emotion-behaviour constructs (i.e., punishment, after-allocation valence and arousal, after-choice valence and arousal, emotional outcome valence and arousal). Each 50×50 correlation matrix represents the inter-correlations between experimental conditions, where each cell ($i,j$) indicates the correlation between condition $i$ and condition $j$ based on their 7-dimensional emotion-behaviour profiles (Supplementary Fig. 2). The correlation matrices were computed using Pearson's correlation coefficient after z-score standardization of all constructs within each group. To further evaluate LLMs' representational similarity to humans, we derived a second-order similarity matrix by correlating vectorized first-order matrices between human and LLM groups (Pearson's $r$), with significance evaluated via Mantel tests (10,000 permutations). The significant exceedance of Pearson's $r$ over the permutation-derived Mantel's $r$ threshold validates these correlations as statistically reliable beyond chance expectations.

**Mediation analyses**

We conducted hierarchical moderated mediation models to test whether after-allocation emotion mediated the relationship between unfair allocations and punishment decisions, using the PROCESS macro (Model 15), BruceR package in RStudio (2022.02.3). Hierarchical linear modeling (HLM) cluster was participant id. Continuous variables were standardized (Z-scores), while binary variables retained their original coding. Mediation effects were estimated using 1,000 Monte Carlo Markov Chain (MCMC) simulations, with standard errors (*SE*) and 95% confidence intervals (CI) reported. Effects were considered statistically significant if their 95% CI did not include zero.

**XGBoost algorithm implementation**

The XGBoost algorithm, a gradient boosting framework based on decision trees[106], was applied to predict punishment behaviour across different groups. The dataset was randomly partitioned into training, validation, and test sets with a 7:1:2 ratio. Specifically, 70% of the data were used to train the classifier, 10% for validation during model development, and 20% were held out as an independent test set for performance evaluation. There were four main features used as independent variables. We first calculated the degree of Unfairness based on the points allocated by Player 1 as one of the original features for the SHAP analysis. This was defined as 30 − 2 × amount of allocation (to Player 2), such that greater unfairness was associated with higher levels of punishment behaviour. The Cost feature was the cost for punishment, with higher costs corresponding to lower levels of punishment. The Emotional Valence and Emotional Arousal features reflected the affective evaluations reported by participants or agents after viewing the allocation. More negative emotions may be associated with punishment behaviour, whereas arousal may have an opposite effect. Importantly, only pre-decision emotional responses were included. The dependent variable was the individual's decision, with punishment coded as 1 and acceptance coded as 0.

The classifier was implemented with the following hyperparameters: **objective** = "binary:logistic", **n_estimators** = 200, **max_depth** = 3, **learning_rate** = 0.05, **subsample** = 0.5, **colsample_bytree** = 0.5, **gamma** = 0.1, **reg_lambda** = 0.1, and **eval_metric** = "logloss". All other hyperparameters were kept at their default settings in the sklearn (version 1.5.1) using Python 3.12. Model performance was assessed on the independent test set using accuracy and area under the receiver operating characteristic curve (AUC). When calculating the AUC, punishment behaviour was coded as 1 and acceptance was coded as 0. To quantify the uncertainty of these estimates, 95% confidence intervals (CIs) were computed using a nonparametric bootstrap procedure with 10,000 iterations.

**SHAP computation**

We computed the SHAP values of different features each trial based on the trained XGBoost classifiers using shap (version 0.48.0) in Python 3.12. Because the classifiers were trained separately for different sample groups, the SHAP values were derived from different models. To ensure comparability across groups, we normalized the SHAP values of each trial to the range of -1 to 1, which enabled us to compare the relative importance of different features between groups[107]. The normalized SHAP was calculated as $\widehat{\phi}_{i,j} = \frac{\phi_{i,j}}{\sum_j |\phi_{i,j}|} \in [-1, 1]$, where $i \in [1, n]$ represents each sample and $j \in [1, 4]$ represents the different SHAP features (i.e., Unfairness, Cost, Emotional Valence and Emotional Arousal). The normalized SHAP values preserve their original signs, reflecting the direction of each feature's influence. The relative importance of features within each group was quantified by summing the absolute values of the normalized SHAP values, thus disregarding the direction of their effects.

**Emotion–Cost Contributions to Punishment**

The emotional contribution to punishment was measured by combining the normalized SHAP values of emotional valence and arousal, whereas the cost contribution was captured by the normalized SHAP value of cost which directly reflects the motivation of self-interest. Considering that the relation between the raw values and normalized SHAP values may differ across groups (for example, in GPT-3.5, the SHAP value of cost increases with the raw cost value, whereas in other models the relation is negative; Supplementary 5.1.2), we first calculated the Pearson's correlation between the raw value and the normalized SHAP value for each participant/agent in each group , which reflect whether a given feature promotes or inhibits punishment behaviour, as well as the degree to which this effect changes with the feature value. These correlations were then used as weights for the absolute values of the normalized SHAP values. When calculating the emotional contribution, we multiplied the valence score by −1 because valence and arousal exerted opposite effects (Supplementary Table 5.1.2). This adjustment ensured that the resulting emotional contribution values were positive, with larger values indicating a stronger promoting effect of emotion on punishment behaviour. In contrast, a larger contribution of cost represents a larger inhibiting effect.

Specifically, for participant $i$, the contribution of emotion and cost to punishment behaviour was calculated as follows:

$$Contribution_{Emotion,i} = (-1) * Corr\left(Valence, \widehat{\phi}_{Valence}\right)_i * \sum_{j=1}^{60} \frac{abs\left(\widehat{\phi}_{Valence,ij}\right)}{60} + Corr\left(Arousal, \widehat{\phi}_{Arousal}\right)_i \sum_{j=1}^{60} \frac{abs\left(\widehat{\phi}_{Arousal,ij}\right)}{60}$$

$$Contribution_{Cost,i} = Corr\left(Cost, \widehat{\phi}_{Cost}\right)_i * \sum_{j=1}^{60} \frac{abs\left(\widehat{\phi}_{Cost,ij}\right)}{60}$$

The $j$ represents each trial and $\widehat{\phi}_{i,j}$ represents normalized SHAP. The functions Corr (x, y) and abs(z) represent Pearson's correlation between x and y, and absolute value of z, respectively.

**Semantic categorization of reasoning content.**

We first defined three theoretically motivated semantic dimensions—emotion, unfairness, and cost—by constructing seed word lists based on prior literature and expert knowledge. A fourth residual category ("other") was included for words not fitting these dimensions.

To build word frequency dictionaries, we compiled all words from chain-of-thought (CoT) outputs and removed stop words, ignoring case distinctions. For human reasoning, the top 180 high-frequency words were selected; for DeepSeek-R1's CoT, the top 1000 words were chosen, covering 85% of total tokens and ensuring representativeness.

Two large language models (Claude 3 and o3-mini) independently classified each high-frequency word into one of the four categories using the curated seed lists as reference (Prompts used see in Supplementary Section 1.6). If their classifications agreed, the category was accepted; if not, words were flagged for manual review with reference to the original CoT context. Inter-model agreement rates were 78.2% for human reasoning and 81.7% for CoT outputs.

For statistical analysis, we constructed contingency tables of category frequencies (emotion, unfairness, cost, other) and compared distributions between humans and the LLM using chi-square tests of independence. Effect sizes were quantified with Cramér's *V*. Standardized residuals were examined to identify which categories contributed most strongly to the observed differences. In addition, word cloud visualizations were generated from the frequency dictionaries to illustrate the semantic emphases of each group.

**Data availability**

All source data are publicly available at https://github.com/liu-h21/LLM-emotion-project.git. The materials used in this study are widely available.

**Code availability**

Data analysis script notebooks are publicly available at https://github.com/liu-h21/LLM-emotion-project.git.